\documentclass[letterpaper]{article} 
\usepackage{aaai25}  
\usepackage{times}  
\usepackage{helvet}  
\usepackage{courier}  
\usepackage[hyphens]{url}  
\usepackage{graphicx} 
\usepackage{natbib}  
\usepackage{caption} 
\usepackage{algorithm}
\usepackage{algorithmic}

\usepackage{newfloat}
\usepackage{listings}
\DeclareCaptionStyle{ruled}{labelfont=normalfont,labelsep=colon,strut=off} 
\floatstyle{ruled}
\newfloat{listing}{tb}{lst}{}
\floatname{listing}{Listing}
\usepackage{booktabs}
\usepackage{multirow}
\usepackage{colortbl}
\usepackage{xcolor}
\definecolor{second}{HTML}{DAE8FC}
\definecolor{best}{HTML}{FFCCC9}
\usepackage{amsmath}
\usepackage{utfsym}
\usepackage{amssymb}

\title{$\textrm{A}^{\textrm{2}}$RNet: Adversarial Attack Resilient Network for \\ Robust Infrared and Visible Image Fusion}
\author{
    Jiawei Li\textsuperscript{\rm 1}\equalcontrib,
    Hongwei Yu\textsuperscript{\rm 1}\equalcontrib,
    Jiansheng Chen\textsuperscript{\rm 1}\thanks{Corresponding author.},
    Xinlong Ding\textsuperscript{\rm 1},
    Jinlong Wang\textsuperscript{\rm 1},\\
    Jinyuan Liu\textsuperscript{\rm 2},
    Bochao Zou\textsuperscript{\rm 1},
    Huimin Ma\textsuperscript{\rm 1}
}
\affiliations{
    \textsuperscript{\rm 1}School of Computer and Communication Engineering, University of Science and Technology Beijing, Beijing, China\\
    \textsuperscript{\rm 2}School of Software Technology, Dalian University of Technology, Dalian, China

    ljw19970218@163.com, yuhongwei22@xs.ustb.edu.cn, jschen@ustb.edu.cn, \{dingxl22, M202320974\}@xs.ustb.edu.cn, atlantis918@hotmail.com, \{zoubochao, mhmpub\}@ustb.edu.cn
}

\usepackage{bibentry}

\begin{document}

\maketitle

\begin{abstract}
Infrared and visible image fusion (IVIF) is a crucial technique for enhancing visual performance by integrating unique information from different modalities into one fused image. Exiting methods pay more attention to conducting fusion with undisturbed data, while overlooking the impact of deliberate interference on the effectiveness of fusion results. To investigate the robustness of fusion models, in this paper, we propose a novel adversarial attack resilient network, called $\textrm{A}^{\textrm{2}}$RNet. Specifically, we develop an adversarial paradigm with an anti-attack loss function to implement adversarial attacks and training. It is constructed based on the intrinsic nature of IVIF and provide a robust foundation for future research advancements. We adopt a Unet as the pipeline with a transformer-based defensive refinement module (DRM) under this paradigm, which guarantees fused image quality in a robust coarse-to-fine manner. Compared to previous works, our method mitigates the adverse effects of adversarial perturbations, consistently maintaining high-fidelity fusion results. Furthermore, the performance of downstream tasks can also be well maintained under adversarial attacks.
\end{abstract}

\begin{links}
	\link{Code}{https://github.com/lok-18/A2RNet}
\end{links}

%

\section{Introduction}
The purpose of infrared and visible image fusion (IVIF) aims to integrate salient information from different sensors for obtaining well-performing fused images, which can alleviate the imaging limitations of a single sensor. The fused image simultaneously contains thermal target information and texture contents from different modalities. In some computer vision tasks, \textit{e.g.}, autonomous driving \cite{sun2022drone} and salient object detection \cite{wang2023interactively}, IVIF technology is applied to assist in achieving more accurate and detailed results.

The primary challenge of this task is how to effectively extract features from different modalities \cite{zhang2021image}. Early traditional methods employ techniques such as wavelet transforms \cite{li1995multisensor} and sparse representation \cite{liu2015general} to perform matrix operations on source images. However, their complex manual adjustment strategies are time-consuming and cumbersome to implement. Recently, deep learning-based methods have gradually replaced traditional approaches \cite{huang2022reconet}. These methods possess powerful feature extraction capabilities to learn salient information from source images \cite{li2022learning, li2023gesenet, liu2023multi}, which has entered an efficient and rapidly evolving stage.

\begin{figure}[]
	\centering
	\includegraphics[width=0.46\textwidth]{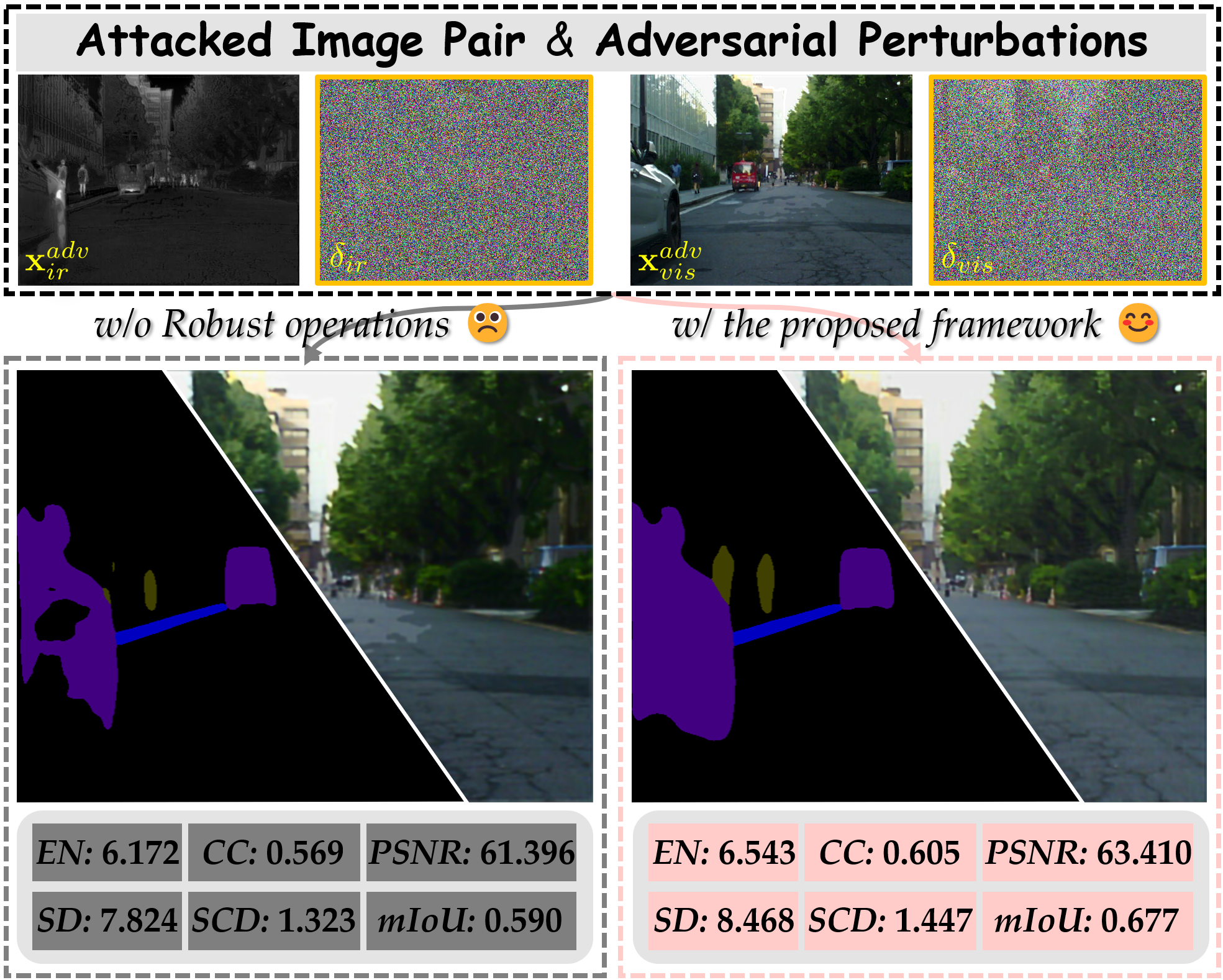}
	\caption{Schematic illustration of different adversarial operations. Clearly, fused images generated by attacked image pairs exhibit superior qualitative, quantitative and downstream task performance when conducting the proposed framework.}
	\label{introduction}	
\end{figure}

In general, existing IVIF methods are based on nondestructive data to construct. They only focus on extracting useful information from source images without considering the potential presence of interference \cite{liu2024task}. In other words, these networks become fragile under adversarial perturbations, which may obtain poor fusion results. Without any robust operations in the network, fused images generated by adversarial examples (AEs) exhibit noticeable artifact regions in Fig.~\ref{introduction}. The segmentation accuracy for some categories also deteriorates accordingly. Some researchers \cite{liu2023paif} have utilized pre-trained segmentation models with adversarial training (AT) to defend fusion networks at a feature-wise level. However, it does not fundamentally design a fusion-oriented paradigm capable of formulating AT. In addition, employing a robust fusion network for resisting perturbations tailored to the IVIF task is also essential for maintaining robustness.

Based on the characteristics of IVIF and existing adversarial research, this paper first develops a novel paradigm with anti-attack loss to achieve AEs and facilitate AT. Then, we propose an adversarial attack resilient network named $\textrm{A}^{\textrm{2}}$RNet for accommodating this paradigm. Specifically, U-Net is employed as the pipeline, where its up-/downsampling operations help filter out noise attacks. To prevent U-Net from overlooking essential features, a transformer-based defensive refinement module (DRM) is implemented in the middle of U-Net, aiming to further refine feature learning and avoid the appearance of noise artifacts. Through the aforementioned network architecture and adversarial strategies, we are able to obtain robust fused images that perform well on both clean and adversarial samples. As depicted in Fig.~\ref{introduction}, the proposed method performs excellent fused images and segmentation results under perturbations. In summary, the main contributions of this paper are as follows:
\begin{itemize}
	\item To achieve a highly-robust fusion framework, we propose an adversarial strategy with a novel anti-attack loss to generate adversarial examples and conduct adversarial training. This approach is rooted in the essence of IVIF and advances the development of adversarial robustness in fusion tasks.
	\item The proposed adversarial attack resilient network ($\textrm{A}^{\textrm{2}}$RNet) uses U-Net as the pipeline for robust feature representation, leveraging its structural characteristics to defend against adversarial perturbations.
	\item Considering that using U-Net may result in texture missing, the defensive refinement module (DRM) is introduced to supplement the extracted information. Furthermore, it enhances the proposed network to resist noise, leading to more refined fusion results.
	\item Extensive experiments demonstrate that our method exhibits stronger robustness under adversarial perturbations. Meanwhile, it also outperforms other comparative methods in downstream task performance.
\end{itemize}

\section{Related Works}

\subsection{Infrared and Visible Image Fusion}
Thanks to the continuous advancements in deep learning technology, infrared and visible image fusion has gradually transitioned from using traditional methods to employing various network architectures\cite{ma2019infrared}, \textit{e.g.}, CNN \cite{cao2023multi}, Transformer \cite{rao2023tgfuse}, GNN \cite{li2023learning} and Diffusion \cite{yue2023dif}. They can effectively extract features from source images and fuse them together, avoiding the hassle of manual adjustment strategies inherent in traditional methods. As a representation, \cite{li2018densefuse} proposed an Auto-Encoder-based fusion network with DenseNet \cite{huang2017densely}, which has been widely adopted in subsequent methods. \cite{liu2021learning} proposed a network structure based on a modified GAN, aiming to address the instability in GAN training while retaining the discriminator to enhance the fidelity of the generated results. In the fusion task, using Transformers to build models allows for a greater focus on global features \cite{ma2022swinfusion}. Recently, the surge in the popularity of generative models has led researchers to incorporate them into IVIF tasks \cite{zhao2023ddfm}. In addition, some key priors or information that help improve fusion results have also been incorporated into networks to assist in feature learning. For instance, \cite{zhao2024image} leveraged large language models, such as GPT3 \cite{brown2020language}, to provide detailed descriptions of source images. Text-IF \cite{yi2024text} also employed a text-guided architecture to construct the fusion network.

\begin{figure*}
	\centering
	\includegraphics[width=0.98\textwidth]{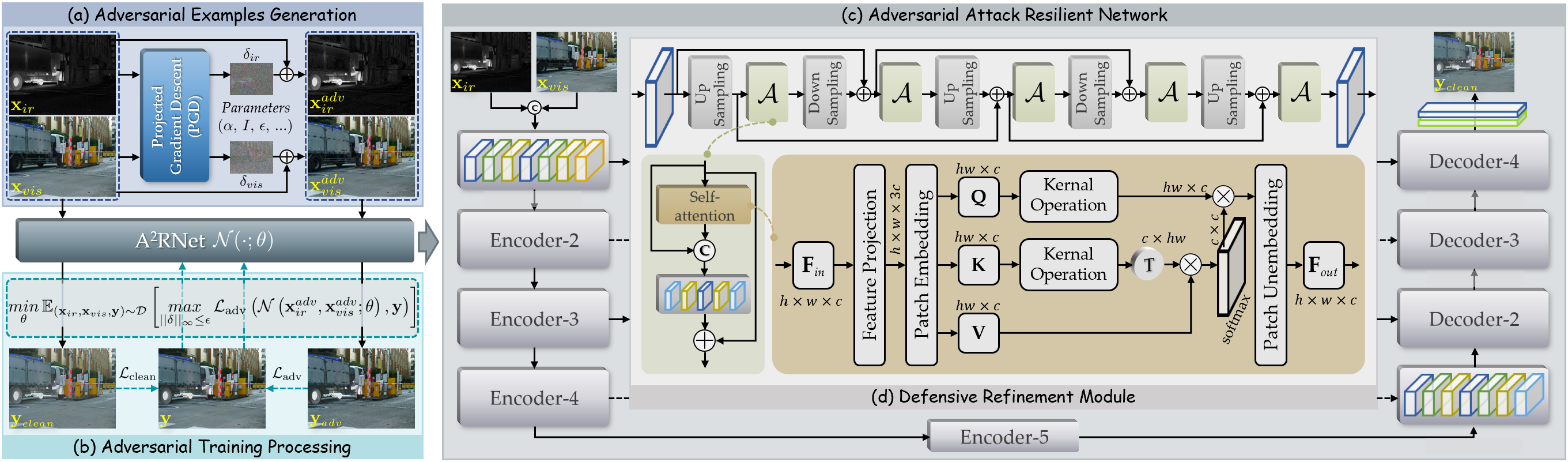}
	\caption{Framework of the proposed $\textrm{A}^{\textrm{2}}$RNet. In specific, (a) and (b) represent the adversarial examples generation and adversarial training processing, respectively. (c) is the adversarial attack resilient network, which contains the defensive refinement module (DRM) as shown in (d).}
	\label{network}	
\end{figure*}

\subsection{Adversarial Attack and Defence}
By adding perturbations to input samples, adversarial attacks aim to mislead deep neural networks (DNN) for producing incorrect outputs. These perturbations are typically difficult for the human visual system (HVS) to detect. As a representative, Fast Gradient Sign Method (FGSM) was proposed by \cite{goodfellow2014explaining}. The process of FGSM can be quantified as:
\begin{equation}
	\delta = \epsilon \cdot \textrm {sign}\left (\bigtriangledown_{\mathbf {x}} \mathcal{L}\left (f\left(\mathbf{x};\theta\right), \mathbf{y}\right ) \right ),
\end{equation}
where $\mathbf{x}$, $\mathbf{y}$ and $f\left(\cdot;\theta\right) $ mean a clean input, its ground truth (GT) and a DNN, respectively. $\mathcal{L}$ is the loss function. $\bigtriangledown$ and $\textrm{sign}(\cdot)$ represent gradients and their direction. $\epsilon$ denote the magnitude of applied perturbation under the specified constraint. With a small step size $\alpha$ for a fixed number of gradient iterations, \cite{madry2017towards} proposed a multi-step optimization variant of FGSM called Projected Gradient Descent (PGD). Similarly, it can be defined as:
\begin{equation}
	\begin{aligned}
		\delta_{k+1} = \delta_{k} + \alpha \cdot \textrm {sign}\left (\bigtriangledown_{ {\mathbf{x}+\delta_{k}}} \mathcal{L}\left (f\left(\mathbf{x} + \delta_{k};\theta\right), \mathbf{y}\right ) \right )  \\
		\mathrm{s.t.} ||\delta||_{p}\leq\epsilon.
	\end{aligned}
	\label{PGD}
\end{equation}
This restriction ensures that $\alpha$ remains within $\epsilon$ around $\mathbf{x}+\delta_{k}$. $p$ represents the norm type. In recent years, an increasing number of adversarial attack methods based on PGD have been proposed in different fields \cite{ding2024transferable, yu2024step}.

In adversarial defence, \cite{yu2022towards} employed an enumeration method to conduct robustness analysis on popular models and loss functions in the deraining task, and combined them into a more robust architecture. As a direct way to improve adversarial robustness, AT involves feeding both clean samples and AEs into the network during training \cite{shafahi2020universal, gokhale2021attribute, jia2022adversarial}. Researchers have quantified the process of AT as a min-max optimization problem \cite{madry2017towards}:
\begin{equation}
	\mathop{\min} \limits_{\theta}\mathbb{E}_{(\mathbf{x}, \mathbf{y})\sim \mathcal{D}} \left[ \mathop{\max}\limits_{||\delta||_{p}\leq\epsilon}\mathcal{L} \left(f(\mathbf{x}+\delta;\theta), \mathbf{y} \right) \right],
	\label{AT}
\end{equation}
where $\mathcal{D}$ denotes the data distribution. For instance, \cite{jiang2024towards} proposed a robust image stitching algorithm with adaptive AT, which enables to resist adversarial attacks and achieve better stitching results. PAIF \cite{liu2023paif} was introudced to conduct IVIF for more robust semantic segmentation. It is the first work related to adversarial robustness in the field of image fusion. Unlike PAIF, our proposed method focuses on the robustness of fused images, then ameliorating the performance of downstream tasks.

\section{Proposed Method}

\subsection{Overview}
For the IVIF task, adversarial attacks seek to disrupt the original representation of source images, causing undesired artifacts and halos in fusion results. To achieve more stable and robust fusion images, we formulate the adversarial attack and training process in IVIF. Note that it is designed based on existing adversarial robustness and the characteristics of IVIF, and exclusively targets the fusion stage. With this formulation, we propose a novel network and a loss function called adversarial attack resilient network ($\textrm{A}^{\textrm{2}}$RNet) and anti-attack loss ($\mathcal{L}_{\mathrm{a}}$) resepctively to facilitate more robust feature learning.

We define $\textrm{A}^{\textrm{2}}$RNet as $\mathcal{N} \left(\cdot; \theta\right)$ parameterized with $\theta$. First, we need to generate AEs with PGD for AT. Different from traditional tasks, the IVIF task involves dual inputs and has no real GT for reference. Therefore, we introduce fair pseudo-labels in conjunction with $\mathcal{L}_{\mathrm{a}}$ to guide the adversarial attacks and training. The inclusion of pseudo-labels ensures that the entire adversarial process is more rational and targeted. After generating the AEs, both clean and attacked results are fed into $\mathcal{N} \left(\cdot; \theta\right)$ simultaneously for training. It is worth noting that our proposed network excels well during AT, which can produce well-performed fusion results under attacks. The specific procedure of the proposed framework is illustrated in Fig.~\ref{network}.

\subsection{Adversarial Attacks and Training in IVIF}
In the initial stage, given a clean image pair $(\mathbf{x}_{ir}, \mathbf{x}_{vis})$ to generate the corresponding AE pair $(\mathbf{x}_{ir}^{adv}, \mathbf{x}_{vis}^{adv}) \leftarrow (\mathbf{x}_{ir} + \delta_{ir}, \mathbf{x}_{vis} + \delta_{vis})$ as illustrated in Fig.~\ref{network}(a). We employ PGD as the AE generator, and its attack process is formulated according to Eq.~\ref{PGD} (taking attacks on infrared images as an example, the generation of 
visible perturbation $\delta^{k+1}_{vis}$ is similar to $\delta^{k+1}_{ir}$) :
\begin{equation}
	\begin{aligned}
		\delta_{ir}^{k+1} = \delta_{ir}^{k} + \alpha \cdot \textrm {sign}\Big(\bigtriangledown_{ {\mathbf{x}_{ir}+\delta_{ir}^{k}}} \mathcal{L}_{\rm{adv}} \big(\mathcal{N}\big(\big(\mathbf{x}_{ir} + \\
		\delta_{ir}^{k}, \mathbf{x}_{vis} + \delta_{vis}^{k}\big);\theta\big), \mathbf{y}\big)\Big) \quad\mathrm{s.t.} ||\delta||_{\infty}\leq\epsilon,
	\end{aligned}
	\label{AE}
\end{equation}
where $\mathcal{L}_{\rm{adv}}$ denotes a part of $\mathcal{L}_{\rm{a}}$, and $\mathbf{y}$ represents the introduced pseudo-label. $l_{\infty}$-norm is chosen to constrain $\epsilon$. In PGD, we conduct $\mathcal{L}_{\rm{adv}}$ for gradient backpropagation and employ pseudo-labels to address the challenge of achieving AEs without GT in IVIF. To ensure relative fairness, we use a common CNN and loss functions, \textit{e.g.}, $l_1$ and SSIM loss, for training and inference to obtain the labels. More details can be found in the supplementary material. Note that not directly using the results of other SOTA methods as pseudo-labels is intended to prevent overfitting. In addition, it can also avoid biasing fusion results towards any particular SOTA method. Therefore, applying this ``moderate" pseudo-labels to the entire process is relatively fair.

To obtain a robust fusion model against attacks, we redesign the AT process based on the characteristics of the IVIF task. Compared to existing methods \cite{liu2023paif}, we focus more on the robustness of fusion, which can first enhance the performance of fused results, then improve the outcomes of downstream tasks. Similarly, Eq.~\ref{AT} is reformulated as: 
\begin{equation}
	{\mathop{min} \limits_{\theta}\mathbb{E}_{(\mathbf{x}_{ir}, \mathbf{x}_{vis}, \mathbf{y})\sim \mathcal{D}}\left[\mathop{max} \limits_{||\delta||_{\infty}\leq\epsilon}\mathcal{L}_{\mathrm{adv}}\left(\mathcal{N}\left(\mathbf{x}_{ir}^{adv}, \mathbf{x}_{vis}^{adv};\mathbf{\theta}\right), \mathbf{y}\right)\right]}
	\label{AT2}
\end{equation}
where $\mathbf{y}$ is also incorporated into AT. As depicted in Fig.~\ref{network}(b), clean $(\mathbf{x}_{ir}, \mathbf{x}_{vis})$ and adversarial examples $(\mathbf{x}_{ir}^{adv}, \mathbf{x}_{vis}^{adv})$ are fed into $\mathcal{N} \left(\cdot; \theta\right)$ to achieve corresponding results separately, \textit{i.e.}, $\mathbf{y}_{clean}$ and $\mathbf{y}_{adv}$. Subsequently, we calculate the loss value with $\mathcal{L}_{\mathrm {clean}}$ and $\mathcal{L}_{\mathrm {adv}}$, which are backpropagated features through the fusion network $\mathcal{N} \left(\cdot; \theta\right)$. The distribution of clean and adversarial examples is ensured to be balanced with this setting, resulting in preventing the robustness bias. 

For the unsupervised IVIF task, generating AEs and performing AT is challenging, so that we employ pseudo-labels to construct a ``supervision" manner. It not only facilitates the effective generation of AEs for AT, but also enhances robustness while ensuring the quality of fusion results. Specifically, the mean squared error (MSE) loss is introduced to estimate the error magnitude between fusion results and labels. Meanwhile, the structural similarity index (SSIM) loss \cite{wang2004image} is used to measure the similarity between them. The basic form of the training loss can be quantified as:
\begin{equation}
	\mathcal{L} = \beta \mathcal{L}_{\mathrm {MSE}} \left(\mathbf{y}^{\prime}, \mathbf{y}\right) + \gamma\left(1-\mathcal{L}_{\mathrm {SSIM}}\left(\mathbf{y}^{\prime}, \mathbf{y}\right)\right),
\end{equation}
where $\beta$ and $\gamma$ are hyperparameters. $\mathbf{y}^{\prime}$ means fusion results. In PGD, we obtain $(\mathbf{x}_{ir}^{adv}, \mathbf{x}_{vis}^{adv})$ by computing $\mathcal{L}_{\mathrm{adv}}$ with $\mathbf{y}_{\mathrm {adv}}$. Note that $\mathcal{L}_{\mathrm{adv}}$ is derived by replacing $\mathbf{y}^{\prime}$ with $\mathbf{y}_{\mathrm {adv}}$. In the stage of AT, $\mathcal{L}_{\mathrm{adv}}$ is used to backpropagate features learned from AEs. Hence, the anti-attack loss for the entire architecture can be expressed as:
\begin{equation}
	\mathcal{L}_{\mathrm{a}} = \mathcal{L}_{\mathrm{clean}} + \mathcal{L}_{\mathrm{adv}},
\end{equation}
where $\mathcal{L}_{\mathrm{clean}}$ denotes the loss incurred during normal training with clean samples.

\subsection{Adversarial Attack Resilient Network}
As essential as adversarial strategies, the design of fusion networks also determines the robustness of fused images. Inspired by techniques such as image restoration \cite{ma2023bilevel, zheng2024u}, Unet-based networks exhibit the excellent capability of feature decoupling. Therefore, we select Unet as the pipeline for our $\textrm{A}^{\textrm{2}}$RNet. However, Unet often experiences some information missing during the decoupling process, which may lead to undesirable artifacts in fusion results. Thanks to the versatility of Unet, we can incorporate flexible modules within it to prevent the aforementioned issues. In short, $\textrm{A}^{\textrm{2}}$RNet is employed to conduct robust feature learning for the IVIF task.

The details of our proposed network are presented in Fig.~\ref{network}(c). In specific, we construct our pipeline by referencing the classical Unet network \cite{ronneberger2015u}. Unlike the common Unet, certain parts of the encoder and decoder, \textit{e.g.}, up/downsampling operation are fine-tuned for better feature extraction. To prevent from missing important details, we propose the defensive refinement module (DRM) in the middle of our Unet pipeline. Considering trade-off, DRM is only connected from Encoder-1/3 to Decoder-4/2. The connections from Encoder-2/4 to Decoder-3/1 conduct the conventional ``copy \& crop" operation.

\begin{algorithm}
	\caption{Adversarial training in $\textrm{A}^{\textrm{2}}$RNet}
	\label{alg}
	\begin{algorithmic}[1]
		\REQUIRE dataset $(\mathbf{x}_{ir}, \mathbf{x}_{vis})\sim\mathcal{D}$, pseudo-labels $\mathbf{y}$, total epoch $T$, network parameters $\theta$
		\FOR {epoch from 1 to $T$}
		\FOR {minibatch $b = 4$}
		\STATE \% \textit{AEs generation with Eq.~\ref{AE}}
		\FOR {$iteration$ form 1 to $I$}
		\STATE $(\delta_{ir}^{i}, \delta_{vis}^{i}) \leftarrow \textrm{PGD}(\alpha, \epsilon, I, \mathbf{x}_{ir}, \mathbf{x}_{vis}, \mathbf{y})$
		\STATE Update $(\delta_{ir}, \delta_{vis}) \leftarrow (\delta_{ir}^{i}, \delta_{vis}^{i})$
		\ENDFOR
		\STATE Generate $(\mathbf{x}_{ir}^{adv}, \mathbf{x}_{vis}^{adv}) \leftarrow (\mathbf{x}_{ir} + \delta_{ir}, \mathbf{x}_{vis} + \delta_{vis})$
		\STATE \% \textit{AT in $\mathcal{N}$ with Eq.~\ref{AT2}}
		\STATE Compute $\mathcal{L}_{\mathrm{clean}}$ on $(\mathbf{x}_{ir}, \mathbf{x}_{vis})$ with $\theta$
		\STATE Compute $\mathcal{L}_{\mathrm{adv}}$ on $(\mathbf{x}_{ir}^{adv}, \mathbf{x}_{vis}^{adv})$ with $\theta$
		\STATE Update $\theta \leftarrow \mathcal{L}_{\mathrm{a}} = \mathcal{L}_{\mathrm{clean}} + \mathcal{L}_{\mathrm{adv}}$
		\ENDFOR
		\ENDFOR
	\end{algorithmic}
\end{algorithm}

\begin{figure*}
	\centering
	\includegraphics[width=0.98\textwidth]{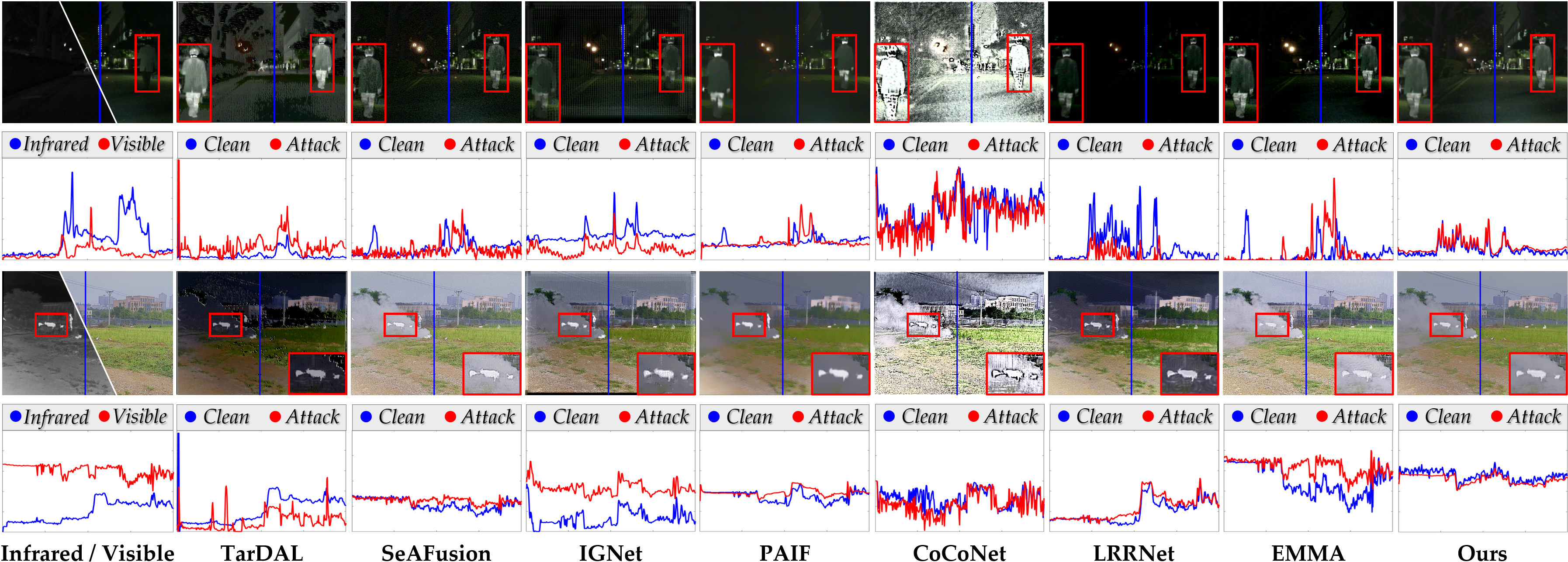}
	\caption{Fusion comparisons with SOTA methods in MFNet and $\textrm{M}^{3}$FD datasets. We apply PGD to clean samples and add perturbations with $\epsilon = 4/255$ to generate AEs. The signal maps are also provided for clean and attack states. The closer the waveform, the stronger the robustness.}
	\label{fusion}	
\end{figure*}

DRM contains five adversarial resistant blocks (ARBs), which leverage features extracted from the Unet for more robust self-attention learning. Multiple sampling and residual operations can also help keep effective contents from source images and filter out attack perturbations during the AT stage. Note that PixelShuffle/Unshuffle \cite{shi2016real} is employed here for up/downsampling. As shown in Fig.~\ref{network}(d), the ARB is illustrated. Similar to the typical transformer architecture, it is also composed of a self-attention layer and a feed-forward layer (including $1 \times 1$ Conv, $3 \times 3$ Conv and LeakyReLU layers). In the self-attention layer, given an input feature $\mathbf{F}_{in} \in \mathbb{R}^{h \times w \times c}$ with height, width and channel dimensions of $h$, $w$ and $c$, feature projection first transforms it into $\mathbf{F}_{in} \in \mathbb{R}^{h \times w \times 3c}$. Next, $\{ {\mathbf{Q}},  {\mathbf{K}},  {\mathbf{V}}\} \in  \mathbb{R}^{hw \times c}$ are obtained through patch embedding. We utilize Mercer's theorem \cite{mercer1909xvi} to contruct a Mercer-based kernel operation for robust feature representation, which reconstructs $\mathbf{Q}$ and $\mathbf{K}$ through the corresponding projection mapping. The Pearson correlation coefficient \cite{cohen2009pearson} is introduced to measure the correlation $r$ between $\mathbf{Q}$ and $\mathbf{K}$, and to validate the kernel operation $m(\cdot)$. It can be defined through Taylor expansion as follow:
\begin{equation}
	K\left(\mathbf{Q, K}\right) = \sum\limits^{\infty}_{i=0}  \dfrac{\left(\mathbf{Q-\bar{Q}}\right)^{2i}}{\sigma^{\frac{1}{2}i}\sqrt{i!}}\dfrac{\left(\mathbf{K-\bar{K}}\right)^{2i}}{\sigma^{\frac{1}{2}i}\sqrt{i!}},
\end{equation}
where $\mathbf{\bar{Q}}$ and $\mathbf{\bar{K}}$ represent the means of $\mathbf{Q}$ and $\mathbf{K}$, respectively. The mapping function is expressed as (taking $\mathbf{Q}$ as an example):
\begin{equation}
	m(\mathbf{Q})=(1,\dfrac{(\mathbf{Q-\bar{Q}})^{2}}{\sigma^{\frac{1}{2}}}, \dfrac{(\mathbf{Q-\bar{Q}})^{4}}{\sigma^{\frac{1}{2}}},\cdot\cdot\cdot,\dfrac{(\mathbf{Q-\bar{Q}})^{2i}}{\sigma^{\frac{1}{2}}}).
\end{equation}
The kernel operation enable to improve the robust representation of self-attention, building a resilient fusion model at the feature-wised level. Moreover, we alter the multiplication order in self-attention by first multiplying $\mathbf{K}$ and $\mathbf{V}$, which reduces the complexity from $O(N^2)$ to $O(N)$ and promotes the efficiency of adversarial training. Therefore, the self-attention matrix is computed as:
\begin{equation}
	\mathrm{Att}(\mathbf{Q}, \mathbf{K}, \mathbf{V}) = \textrm{softmax} \left(\dfrac{\mathbf{K} \cdot m(\mathbf{V})^{\mathrm{T}}}{\sqrt{d_{s}}}\right)m(\mathbf{\mathbf{Q}}),
\end{equation}
where $d_s$ and $\mathrm T$ are the scaling factor and transpose operation. Finally, $\mathbf{F}_{out} \in \mathbb{R}^{h \times w \times c}$ is achieved through patch unembedding and fed into Decoder-4/2. We detail the entire process in Algorithm.\ref{alg}, including the steps for adversarial example generation, robust feature learning, and adversarial training.

\section{Experiments}

\subsection{Experimental Setup Details}

Before adversarial training, we first need to generate adversarial examples by using PGD. Specifically, a moderate perturbation is set with a total iteration $I$ of 3, a perturbation strength $\epsilon$ of 4/255, and a step size $\alpha$ of 1/255. This configuration helps to avoid excessive time spent on getting adversarial examples. $l_{\infty}$-norm constrains perturbations. During the adversarial training phase, the Adam optimizer is chosen to adjust $\theta$ with a 0.001 learning rate. We set the batch size and total epochs to 4 and 50, respectively. In $\mathcal{L}$, $\beta$ and $\gamma$ are 100. The number of clean and adversarial examples is kept at 1:1 for balance. $\textrm{M}^{3}$FD \cite{liu2022target} and MFNet \cite{ha2017mfnet} datasets are introduced for training and testing. In the adversarial inference stage, we set $I$ to 20 with unchanged $\epsilon$ and $\alpha$ to generate AEs and use them to obtain fused images. It is noticed that all experiments are conducted on an Intel(R) Xeon(R) Gold 6271C CPU and a NVIDIA Tesla A100 GPU with PyTorch.

\begin{figure}[h]
	\centering
	\includegraphics[width=0.46\textwidth]{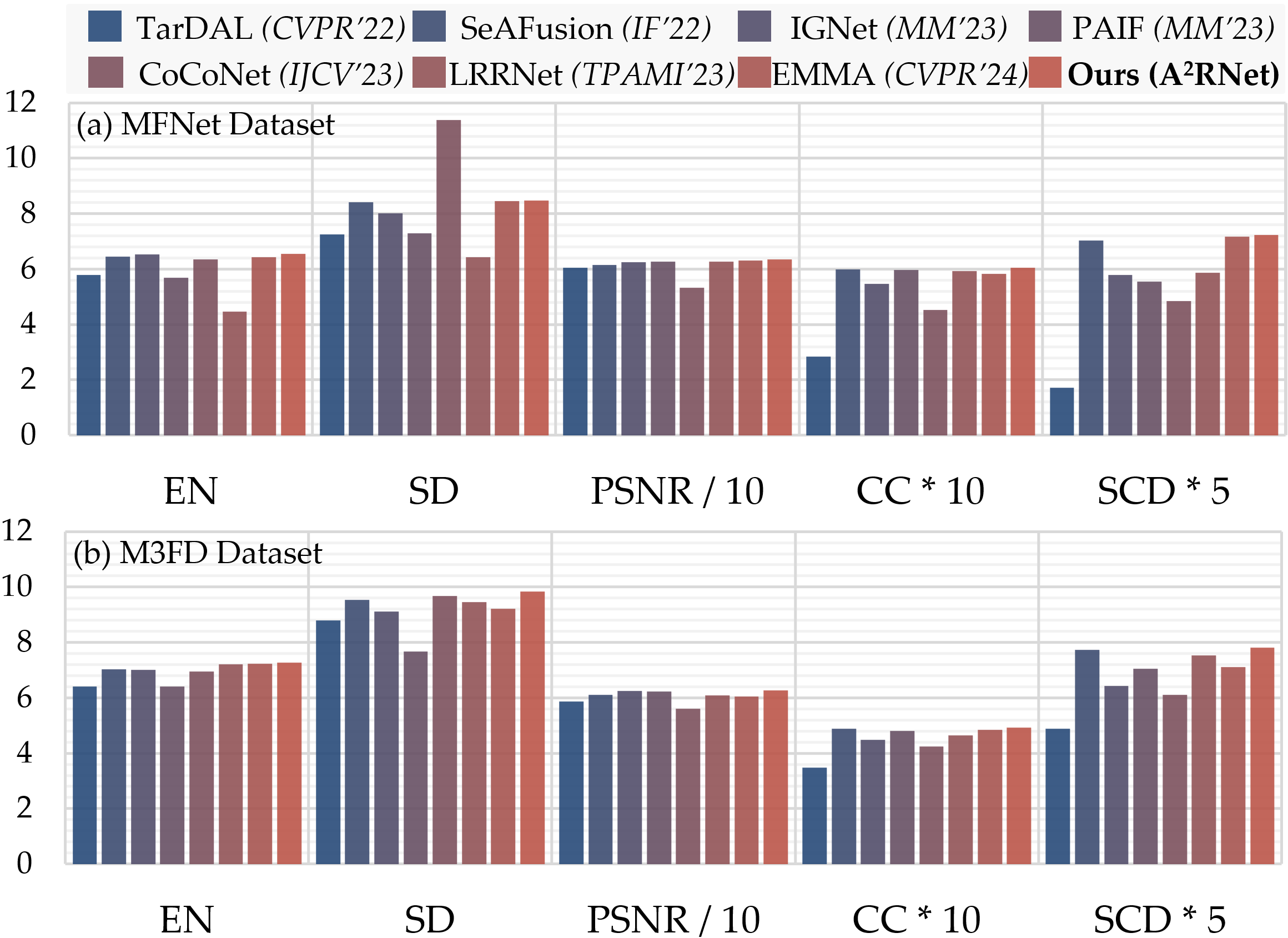}
	\caption{Bar charts of the fusion comparison metrics. For better visualization, we have scaled the values of certain metrics.}
	\label{fusion quantitative}	
\end{figure}

\begin{figure*}[h]
	\centering
	\includegraphics[width=1\textwidth]{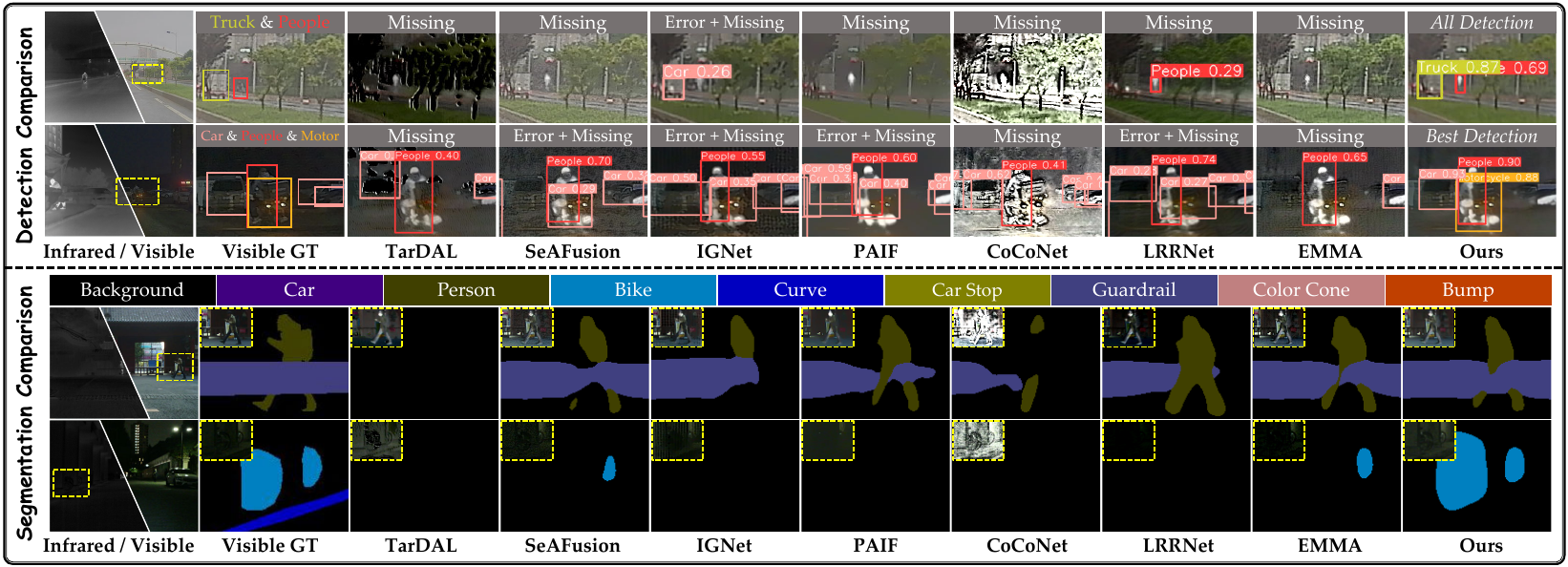}
	\caption{Detection and segmentation comparisons of fused images. Under adversarial conditions, our method yields better performance in downstream tasks.}
	\label{det-seg}	
\end{figure*}

\begin{table*}[h]
	\centering
	\renewcommand\tabcolsep{4pt} 
	\begin{tabular}{c|cccccccc}
		\toprule
		\multirow{2}{*}{\textbf{Metric}} & \multicolumn{8}{c}{\textbf{Method}}                                   
		\\
		& TarDAl & SeAFusion & IGNet & PAIF  & CoCoNet & LRRNet & EMMA  & Ours  
		\\ 
		\midrule
		$\mathrm{mAP@.5}$ & 0.462$_{\scriptscriptstyle \downarrow0.161}$  & {\colorbox{second}{0.732}$_{\scriptscriptstyle \downarrow0.042}$} & 0.557$_{\scriptscriptstyle \downarrow0.154}$ & 0.715$_{\scriptscriptstyle \downarrow0.046}$ & 0.311$_{\scriptscriptstyle \downarrow0.093}$ & 0.704$_{\scriptscriptstyle \downarrow0.032}$  & 0.696$_{\scriptscriptstyle \downarrow0.038}$ & \colorbox{best}{0.781}$_{\scriptscriptstyle \downarrow0.024}$ 
		\\
		$\mathrm{mIoU}$& 0.415$_{\scriptscriptstyle \downarrow0.069}$ & 0.618$_{\scriptscriptstyle \downarrow0.078}$ & 0.514$_{\scriptscriptstyle \downarrow0.184}$ & 0.578$_{\scriptscriptstyle \downarrow0.134}$ & 0.409$_{\scriptscriptstyle \downarrow0.080}$ & 0.623$_{\scriptscriptstyle \downarrow0.027}$ & $\colorbox{second}{0.649}_{\scriptscriptstyle \downarrow0.086}$ & $\colorbox{best}{0.677}_{\scriptscriptstyle \downarrow0.068}$ 
		\\ 
		\bottomrule
	\end{tabular}
	\caption{Quantitative results of detection (mAP@.5) and segmentation (mIoU). \colorbox{best}{Red} and \colorbox{second}{blue} denote the optimal and suboptimal results, respectively. The subscripts indicate the change compared adversarial conditions with the clean.}
	\label{detection and segmentation}
\end{table*}

To demonstrate the superiority of our method, we conduct comparisons in both qualitative and quantitative results. Seven SOTA methods are selected for comprehensive comparison, including TarDAL \cite{liu2022target}, SeAFusion \cite{tang2022image}, IGNet \cite{li2023learning}, PAIF \cite{liu2023paif}, CoCoNet \cite{liu2024coconet}, LRRNet \cite{li2023lrrnet} and EMMA \cite{zhao2024equivariant}. Except for PAIF, none of the other methods have investigated adversarial robustness. To ensure fairness, we apply the same adversarial settings as our method to the open-source code of these approaches. In the quantitative comparison, we choose Entropy (EN), Standard Deviation (SD), Peak Signal-to-Noise Ratio (PSNR), Correlation Coefficient (CC) \cite{shah2013multifocus} and the Sum of the Correlations of Differences (SCD) \cite{aslantas2015new}. Higher values indicate better image performance. In the comparison of downstream tasks, mean average precision (mAP@.5) and mean intersection over union (mIoU) are used to evaluate detection and segmentation, respectively. The experimental details of the downstream tasks are provided in the supplementary materials.

\subsection{Comparison Results}

\subsubsection{Comparison of Fusion Results} Fig.~\ref{fusion} presents the qualitative comparison results of our method and other SOTA approaches under adversarial attacks. The performance of TarDAL and CoCoNet is noticeably inconsistent with HVS, exhibiting evident attacked regions such as noisy spots on the ground (first set) and color distortions on the grass (second set). SeAFusion, IGNet and EMMA all exhibit varying degrees of noisy textures, which significantly impact the visual quality. For instance, in the magnified patch of the person in the first set, the details on the clothing are not as smooth as the clean image. Although LRRNet does not contain excessive noise, it compromises on brightness. In the first set of results, the details on the road are not clearly observed, meanwhile in the second set, the sky appears overly dark, which does not align with realistic natural scenes. It indicates that adversarial examples disrupt the balance of information extraction in the original network. In addition, we also provide signal maps for each method under different inference conditions, \textit{i.e.}, clean and attack. Since PAIF incorporates robustness operations, it can withstand certain levels of attacks. However, it exhibits an over-smooth phenomenon or even blurry texture, which is an undesirable outcome. Thanks to the effective adversarial training strategy and network architecture in our method, we achieve more stable and robust fusion results. They not only capture the desired detailed features but also show minimal differences compared to the clean results. As shown in Fig.~\ref{fusion quantitative}, we present the quantitative comparison results with bar charts. It can also prove that our method obtains superior results.

\subsubsection{Comparison of Detection Results}
The fusion results obtained from AEs also lead to some changes in downstream tasks. In the detection task, we provide two sets of enlarged patches on the $\textrm{M}^{3}$FD dataset as shown in Fig.~\ref{det-seg}. In the first set, the patch contains a truck and a person. Due to the poor visual quality of TarDAL and CoCoNet, the detector fails to identify any targets. SeAFusion and EMMA are affected by noisy spots, misleading the detector into making incorrect judgments. As the edge features of each target are not distinct in PAIF, the detection network is unable to capture the necessary information for detection. The results from IGNet and LRRNet also exhibit some errors or missing detections with low confidence. However, the proposed method can accurately detect all targets with high confidence. There are more objects to detect in second comparisons. Compared to other methods, we obtain the best detection results. In the quantitative comparison, Table.~\ref{detection and segmentation} presents the mAP@.5 for all methods. The subscripts indicate the difference between clean and attack detection results. From the subscripts of mAP@.5, it can be seen that our results not only achieve the highest scores but also maintain the smallest difference compared to the clean. The specific AP@.5 values for each category are provided in the supplementary materials.

\subsubsection{Comparison of Segmentation Results}
Similarly, we present the qualitative and quantitative segmentation results on the MFNet dataset in Fig.~\ref{det-seg} and Table.~\ref{detection and segmentation}, respectively. The detailed AP@.5 values for each category are presented in the supplementary materials. In the daytime scenes, fused images with perturbations exhibit inaccurate regions in the segmentation results. For instance, SeAFusion, IGNet and CoCoNet perform poorly on the ``Person" category. LRRNet shows undesirable results at the boundary between ``Guardrail" and ``Person", struggling to accurately differentiate foreground information. Although EMMA achieves relatively good segmentation results, it still falls short compared to ours. It can be substantiated by both the mIoU values and ground truth. In the nighttime scenes, only SeAFusion and EMMA are able to segment parts of the ``Bike" category. TarDAL fails to depict any meaningful information in all results. However, our method achieves superior visual performance and quantitative metrics compared to all SOTA methods, which proves that our fusion results can maintain a robust state in the segmentation task.

\subsection{Ablation Analysis}

\begin{figure}
	\centering
	\includegraphics[width=0.46\textwidth]{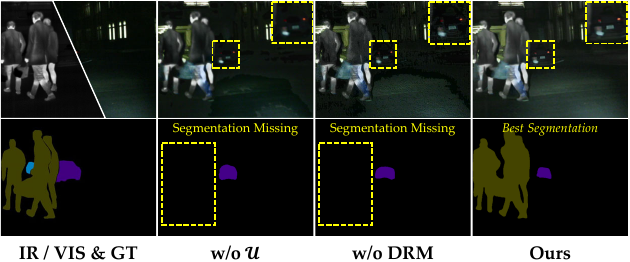}
	\caption{Ablation analysis in different modules.}
	\label{ablation module}	
\end{figure}

\begin{table}[]
	\centering
	\renewcommand\tabcolsep{2pt} 
	\begin{tabular}{ccc|cccccc}
		\toprule
		& & & \multicolumn{6}{c}{\textbf{Dataset:MFNet}}              
		\\
		\multirow{-2}{*}{\textbf{Model}} & \multirow{-2}{*}{$\mathcal{U}$} & \multirow{-2}{*}{DRM} & EN & SD & PSNR & CC & SCD & mIoU  
		\\ 
		\midrule
		M1 & \usym{2717} & \usym{2714} & 5.231 & 6.942 & 60.673 & 0.593 & 1.403 & 0.594 
		\\
		M2 & \usym{2714} & \usym{2717} & 5.742 & 7.586 & 63.157 & 0.587 & 1.325 & 0.669 
		\\
		\rowcolor[HTML]{FFCCC9} 
		M3 & \usym{2714} & \usym{2714} & 6.543 & 8.468 & 63.410 & 0.605 & 1.447 & 0.745 
		\\ 
		\bottomrule
	\end{tabular}
	\caption{Quantitative ablation results of different modules.}
	\label{ablation module quantitative}
\end{table}

\subsubsection{Analysis of Modules}
In the proposed method, the synergy between the Unet pipeline and DRM is a key reason for maintaining the robustness of network. We conduct ablation studies by progressively disabling these two modules. The corresponding qualitative and quantitative comparison results are presented in Fig.~\ref{ablation module} and Table.~\ref{ablation module quantitative}. Note that we introduce $\mathcal{U}$ to represent Unet. Without $\mathcal{U}$, undesirable patches appear in the fusion results, most notably on the ground. Omitting DRM may cause prominent noisy areas, particularly at the edges of cars and windows. The segmentation results of them completely fail to capture the ``Person" category. Obviously, our results not only avoid artifacts but also prevent the occurrence of objectionable noise. It indicates that using $\mathcal{U}$ allows the architecture to filter out most perturbations. Additionally, DRM can further resist noise attacks and refine feature expressions to achieve robust fusion images. In the segmentation performance, our results are the most similar to the ground truth. The quantitative results in Table.~\ref{ablation module quantitative} also demonstrate that our architecture plays a positive role in constructing a robust IVIF network.

\begin{figure}
	\centering
	\includegraphics[width=0.46\textwidth]{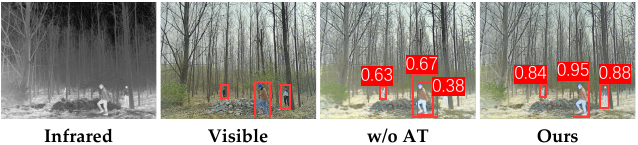}
	\caption{Ablation analysis of adversarial training.}
	\label{ablation advtraining}	
\end{figure}

\subsubsection{Analysis of Adversarial Training}
Apart from the contribution of modules, AT is also significant in enhancing the robustness of our network. We keep the original model and experimental settings unchanged with different training strategies, \textit{i.e.}, AT \textit{vs.}  non-AT. When the network conducts without AT and subjected to attacks, we can observe in Fig.~\ref{ablation advtraining} that the fusion results exhibit noticeable blurriness. Moreover, not all targets in the scene are detected well. It proves that the robustness of the network without AT is still compromised by perturbations, leading to less-than-ideal fusion quality and performance in downstream tasks. In contrast, we can achieve more robust and stable results with the proposed adversarial strategy designed for the fusion task. More details and targets are able to be observed and detected in our method. The metrics with AT are also significantly higher than those without, which is shown in Fig.~\ref{ablation advtraining}.

\begin{figure}
	\centering
	\includegraphics[width=0.46\textwidth]{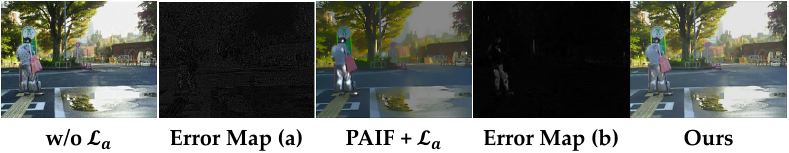}
	\caption{Ablation analysis of $\mathcal{L}_{\mathrm a}$. Error maps between original and ablation results are given to observe differences.}
	\label{ablation loss}	
\end{figure}

\subsubsection{Analysis of Loss Function}
Another ablation study is also to investigate whether the robustness of network would degrade if $\mathcal{L}_{\mathrm a}$ is not used. In Fig.~\ref{ablation loss}, we present the ablation results without $\mathcal{L}_{\mathrm a}$. Instead, we employ the weighted average of source images as $\mathbf{y}$ and common loss functions \cite{tang2022image} into Eq.~\ref{AE} and~\ref{AT2} for ablation experiments. From the error map (a), the fused image exhibits noticeable noise compared to the clean image. In addition, we embed $\mathbf{y}$ and $\mathcal{L}_{\mathrm a}$ into PAIF and retrain the model to verify whether they can also enhance the existing robustness. Apart from slight changes in luminance, the texture features and details of the targets do not improve. Therefore, it can be concluded that $\mathcal{L}_{\mathrm a}$ is not a plug-and-play loss function, which needs to work with the proposed network to achieve more robust representations.

\section{Conclusion}
This paper proposed a robust method for infrared and visible image fusion that is designed to endure adversarial disturbances. Based on the intrinsic nature of the fusion task, we conducted the adversarial attack and training processes by using the proposed anti-attack loss. During the training phase, we employed a Unet-based architecture and a transformer-based defensive refinement module to equip the network with a coarse-to-fine noise filtering capability. The defensive refinement modelue also comliemented missing features to refine textures of fused images. Compared to existing methods, $\textrm{A}^{\textrm{2}}$RNet demonstrates strong resilience against perturbations. Moreover, it maintains a high level of performance in downstream tasks under attack. In future works, we should also focus on the robustness of the IVIF task from data to enhance the performance of fused images.

\section*{Acknowledgements}
This work was supported by the National Science and Technology Major Project (2022ZD0117902), and by the National Natural Science Foundation of China (62376024, 62206015) and the Fundamental Research Funds for the Central Universities (FRF-TP-22-043A1).

\bibliography{Li}

\end{document}